\documentclass[11pt,a4paper]{article}
\usepackage[hyperref]{acl2021}
\usepackage{times}
\usepackage{latexsym}
\usepackage{url}
\usepackage{times}
\usepackage{soul}
\usepackage{url}
\usepackage[utf8]{inputenc}
\usepackage{graphicx}
\usepackage{amsmath}
\usepackage{amsthm}
\usepackage{booktabs}
\usepackage{algorithm}
\usepackage{algorithmic}
\usepackage{color}
\usepackage{tablefootnote}

\newcommand{\laco}{\textsc{Laco}}

\newcommand*\samethanks[1][\value{footnote}]{\footnotemark[#1]}
\newcommand\blfootnote[1]{%
\begingroup
\renewcommand\thefootnote{}\footnote{#1}%
\addtocounter{footnote}{-1}%
\endgroup
}
\usepackage{microtype}
\title{Enhancing Label Correlation Feedback in Multi-Label Text Classification via Multi-Task Learning}

\aclfinalcopy
\author{Ximing Zhang$^1$\thanks{\ \ Equal contribution.}$\ \ {}^\dagger$ , Qian-Wen Zhang$^2$\samethanks, Zhao Yan$^2$, Ruifang Liu$^1$, Yunbo Cao$^2$\\
$^1$Beijing University of Posts and Telecommunications, Beijing 100876, China\\
$^2$Tencent Cloud Xiaowei, Beijing 100080, China\\
ximingzhang@bupt.edu.cn, cowenzhang@tencent.com, zhaoyan@tencent.com,  \\
lrf@bupt.edu.cn,  yunbocao@tencent.com
}
\date{}

\begin{document}
\maketitle
\begin{abstract}
\blfootnote{$^\dagger$Work done during an internship at Tencent.}In multi-label text classification (MLTC), each given document is associated with a set of correlated labels. To capture label correlations, previous classifier-chain and sequence-to-sequence models transform MLTC to a sequence prediction task. 
However, they tend to suffer from label order dependency, label combination over-fitting and error propagation problems.
To address these problems, we introduce a novel approach with multi-task learning to enhance label correlation feedback.
We first utilize a joint embedding (JE) mechanism to obtain the text and label representation simultaneously.
In MLTC task, a document-label cross attention (CA) mechanism is adopted to generate a more discriminative document representation. 
Furthermore, we propose two auxiliary label co-occurrence prediction tasks to enhance label correlation learning:
1) Pairwise Label Co-occurrence Prediction (PLCP), and 2) Conditional Label Co-occurrence Prediction (CLCP).
Experimental results on AAPD and RCV1-V2 datasets show that our method outperforms competitive baselines by a large margin.
We analyze low-frequency label performance, label dependency, label combination diversity and coverage speed to show the effectiveness of our proposed method on label correlation learning. Our code is available at {\url{https://github.com/EiraZhang/LACO}}.
% \footnote{ Our code is available at https://github.com/EiraZhang/LACO}

\end{abstract}
\section{Introduction}
Multi-label text classification (MLTC) is an important natural language processing task with applications in text categorization, information retrieval, web mining, and many other real-world scenarios \cite{zhang2013review,liu2020emerging}. 
In MLTC, each given document is associated with a set of labels which are often related statistically and semantically. 
Label correlations should be sufficiently utilized to build multi-label classification models with strong generalization performance \cite{tsoumakas2009mining,gibaja2015tutorial}.  
In particular, learning the dependencies between labels might be helpful in modeling the low-frequency labels, because real-world classification problems tend to exhibit \emph{long-tail} label distribution, where low-frequency labels are associated with only a few instances and are difficult to learn \cite{menon2020long}.

Previous sequence-to-sequence (Seq2Seq) based methods \cite{nam2017maximizing,yang2018sgm} have been shown to have a powerful ability to capture label correlations with using the current hidden state of the model and the prefix label predictions.
However, exposure bias phenomenon \cite{bengio2015scheduled} may cause the models overfit to the frequent label sequence in training set, thus lead to several problems.
First, 
Seq2Seq-based methods heavily rely on a predefined ordering of labels and perform sensitively to the label order \cite{vinyals2015order,yang2019deep,qin2019adapting}. 
Actually, labels are essentially an order-independent set in the MLTC task.
Second, the Seq2Seq-based methods suffer from low generalization ability problem since they tend to overfit the label combinations in the training set and have difficulty to generate the unseen label combination.
Third, Seq2Seq-based methods rely on the previous potentially incorrect prediction results. The errors may propagate during the inference stage where true previous target labels are unavailable and are thus replaced by labels generated by the model itself.

To circumvent the potential issues mentioned above, 
we introduce a multi-task learning based approach that does not rely on Seq2Seq architecture. The approach contains a shared encoder, a
MLTC task specific module and a label correlation enhancing module.
In the shared parameter layers, we introduce a joint embedding (JE) mechanism which takes advantage of a transformer-based encoder to obtain document and label representation jointly.
Correlations among labels are learned implicitly through the self-attention mechanism, which is different from previous label embedding methods \cite{wang2018joint,xiao2019label} that treat labels independently.
In MLTC task specific module, we generate the label-specific document representation by the document-label cross attention (CA) mechanism, which retains discriminatory information. The shared encoder and the MLTC task specific module form the basic model called \textbf{{\laco}}, i.e. LAbel COrrelation aware multi-label text classification.

The co-occurrence relationship among labels is one of the important signal that can reflect label correlations explicitly, which can be obtained without additional manual annotation.
%  by joint training with the MLTC task
In label correlation enhancing module, we propose two label co-occurrence prediction tasks, which are jointly trained with the MLTC task.
The one is the \emph{Pairwise Label Co-occurrence Prediction} (PLCP) task for capturing second-order label correlations through the two-by-two combinations to distinguish whether they appear together in the set of relevant labels.
The other one is the \emph{Conditional Label Co-occurrence Prediction} (CLCP) task for capturing high-order label correlations through a given partial relevant label set to predict the relevance of other unknown labels. 

We conduct experiments on AAPD and RCV1-V2 datasets, and show that our method outperforms competitive baselines by a large margin.
Comprehensive experimental results are provided to analysis low-frequency label performance, label dependency, label combination diversity and coverage speed, which are essential to measure the ability of label correlation learning. We highlight our contributions as follows: 
\begin{enumerate}
\item We propose a novel and effective approach for MLTC, which not only sufficiently learns the features of documents and labels through the joint space, but also reinforces correlations through multi-task design without depending on the label order.
\item We propose two feasible tasks (PLCP and CLCP) to enhance the feedback of label correlations, which is beneficial to help induce the multi-label predictive model with strong generalization performance.
\item We compare our approach with competitive baseline models on two multi-label classification datasets and systematically demonstrate the superiority of the proposed models.
\end{enumerate}

\section{Related Work}\label{related}
Our work mainly relates to two fields of MLTC task: label correlation learning and document representation learning.
\subsection{Label Correlation Learning}
For MLTC task, a simple but widely used method is binary relevance (BR) \cite{boutella2004learning}, which decomposes the MLC task into multiple independent binary classification problem without considering the correlations between labels.

To capture label correlations, label powerset (LP) \cite{tsoumakas2007multi} take MLTC task as a multi-class classification problem by training a classifier on all unique label combinations. 
% Obviously, this method cannot predict label cominations that have not appeared in the training set.
Classifier Chains (CC) based method \cite{read2011classifier} 
exploits the chain rule and predictions from the previous classifiers as input.
Seq2Seq architectures are proposed to transform MLTC into a label sequence generation problem by encoding input text sequences and decoding labels sequentially \cite{nam2017maximizing}.
However, both CC and Seq2Seq-based methods heavily rely on a predefined ordering of labels and perform sensitively to the label order.
To tackle the label order dependency problem, various methods have been explored: by sorting heuristically \cite{yang2018sgm}, by dynamic programming \cite{liu2015optimality}, by reinforcement learning \cite{yang2019deep}, by multi-task learning \cite{order-free:aaai2020,Seq2Seq+br}.
Different from these works, our method learns the label correlations through a non-Seq2Seq-based approach without suffering the above mentioned problems.

More recently, researchers have proposed a variety of label correlation modeling methods for MLTC that are not based on Seq2Seq architecture.
\citet{multi-reasoner:nju2020} propose a multi-label reasoner mechanism that employs multiple rounds of predictions,
and relies on predicting multiple rounds of results to ensemble or determine a proper order, which is computationally expensive.
CorNetBertXML \cite{correlation-network:kdd2020} utilizes BERT \cite{devlin2019bert} to obtain the joint representation of text and all candidate labels and extra exponential linear units (ELU) at the prediction layer to make use of label correlation knowledge. 
Different from the above works, we exploit extra label co-occurrence prediction tasks to explicitly model the label correlations in a multi-task framework.
\subsection{Document Representation Learning }
Text representation plays a significant role in text classification tasks. It is crucial to extract essential hand-crafted features for early models \cite{joachims1998text}. 
Deep neural network based MLTC models have achieved great success such as CNN \cite{kurata2016improved,liu2017deep}, RNN \cite{liu2016recurrent}, CNN-RNN \cite{chen2017ensemble,lai2015recurrent}, attention mechanism \cite{yang2016hierarchical,you2018attentionxml,adhikari2019docbert} and etc.
\cite{devlin2019bert} is an important turning point in the development of text classification task and it works by generating contextualized word vectors using Transformer.
The reason why deep learning methods have become so popular is their ability to learn sophisticated semantic representations from text, which are much richer than hand-crafted features\cite{guo2020label}.
However, these methods tend to ignore the semantics of labels while focusing only on the representation of the document.

Recently, label embedding is considered to improve multi-label text classification tasks.
% \cite{liu2017deep, pappas2019gile}.
\cite{liu2017deep} is the first DNN-based multi-label embedding method that seeks a deep latent space to jointly embed the instances and labels.
LEAM \cite{wang2018joint} applies label embedding in text classification, which obtains each label’s embedding by its corresponding text descriptions.
LSAN \cite{xiao2019label} makes use of document content and label text to learn the label-specific document representation with the aid of self-attention and label-attention mechanisms.
Our work differs from these works in that the goal of our work is to consider not only the relevance between the document and labels but also the correlations between labels.
\section{Methodology}
The framework of {\laco} is shown in Figure \ref{model_3task}. The lower layers are shared across all tasks, while the top layers are task-specific. In this section, we first introduce the standard formal definition of MLTC. After that, we present the detailed technical implementation of {\laco}. 
\subsection{Problem Formulation}
\label{Problem Formulation}
Multi-label task studies the classification problem where each single instance is associated with a set of labels simultaneously.
Given a training set $S = \{{(D_i}, Y_i^+)|1\leq i\leq N\}$ of multi-label text classification data, $D_i$ is the text sequence and $Y_i^+$ is its corresponding labels.
Specifically, a text sequence $D$ of length $m$ is composed of word tokens $D = \{x_1, x_2, ..., x_m \}$, and $Y = \{y_1, y_2, ..., y_n\}$ denote the label space consisting of $n$ class labels.
The aim of MLTC is to learn a predictive function $f: D \rightarrow 2^{Y}$ to predict the associated label set for the unseen text. For such, the model must optimize a loss function which ensures that the relevant and irrelevant labels of each training text are predicted with minimal misclassification.
\subsection{Document-Label Joint Embedding (JE)}
% \subsection{Label Correlation Aware Network}
\label{JE}
\begin{figure}[t]
	\small
	  \centering
	   \includegraphics[width=7.5cm]{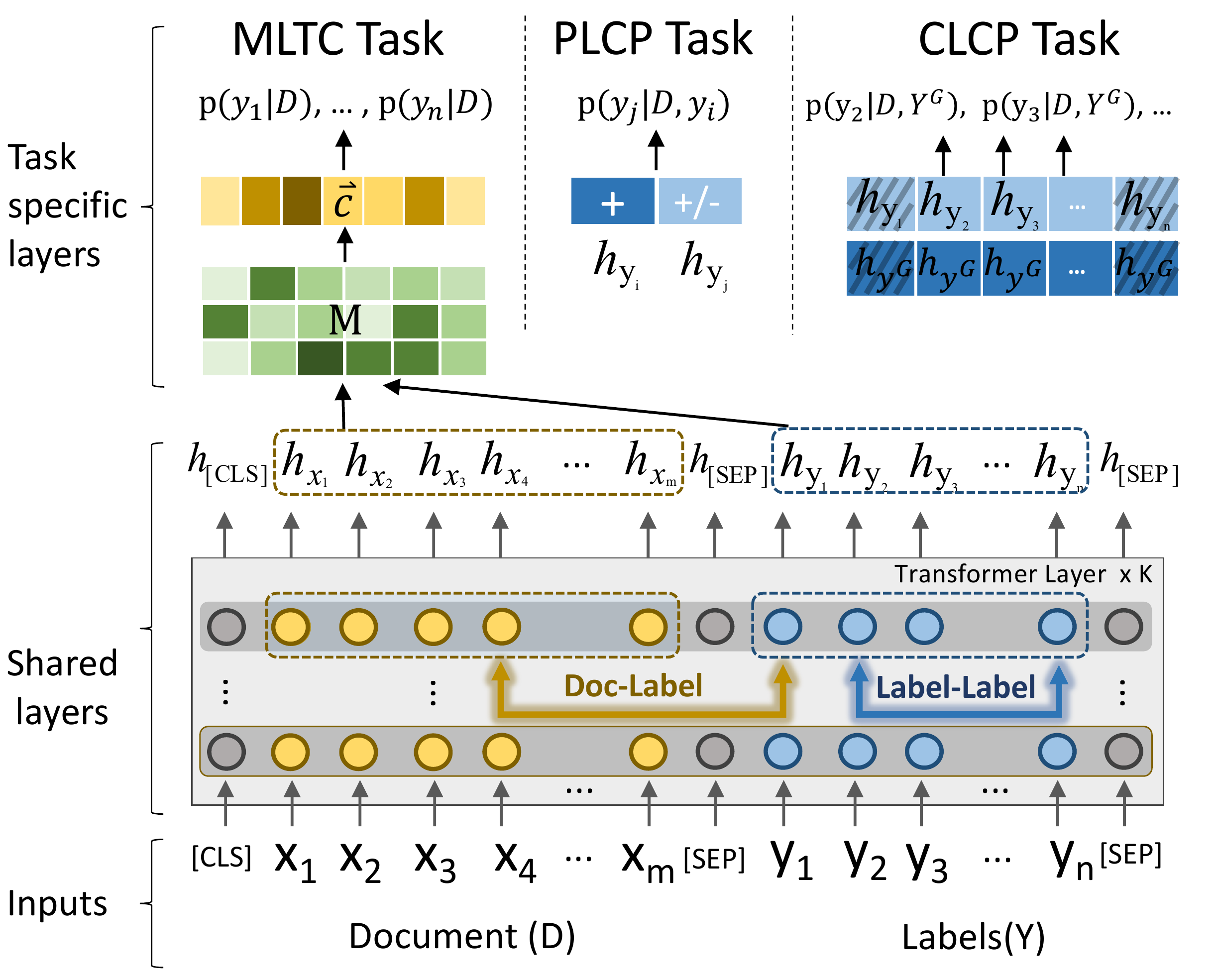}
	   \caption{The framework of our proposed approach. Note that the shaded square in the CLCP task is the embedding of given labels, and $+$, $-$ represent related label and unrelated label respectively. }\label{model_3task}
\end{figure}
Following BERT \cite{devlin2019bert}, the first token is always the [CLS] token. The output vector corresponding to the [CLS] token aggregates the features of the whole document and can be used for classification.
Different from this habitual operation, we propose a novel input structure to directly use label information in constructing the token-level representations.

As shown in Figure \ref{model_3task}, the inputs are packed by a sequence pair $(D, Y)$, we separate the text sequence $D$ and the label sequence $Y$ with a special token [SEP]. Note that the label sequence is to concatenate all label tokens. The shared layers map the inputs into a sequence of embedding vectors, one for each token, called token-level representations. Formally, let $\{[CLS], x_1, ..., x_m, [SEP], y_1, ..., y_n, [SEP]\}$ be the input sequence of the encoder, we obtain the output contextualized token-level representations $\{h_{[CLS]}, h_{x_1},..., h_{x_m}, h_{[SEP]}, h_{y_1}, ..., h_{y_n}, h_{[SEP]}\}$.
The input structure is designed to guarantee that words and labels are embedded together in the same space. 
With the joint embedding mechanism, our model could pay more attention to two facets: 1) The correlations between document and labels. Different document have different influences on a specific label, while the same document fragment may affect multiple labels. 2) The correlations among labels. The semantic information of labels is interrelated, and label co-occurrence indicates strong semantic correlations between them.  
\subsection{Multi-Label Text Classification}
\label{model}
In this subsection, we introduce the MLTC task specific module, including \emph{Document-Label Cross Attention} (CA) and \emph{Label Predication}.
\subsubsection{Document-Label Cross Attention (CA)}
To explicitly model the semantic relationship between each word and label token, we measure the compatibility of label-word pairs via dot product:
\begin{equation}
M = H_D H_Y^T
\end{equation}
% $H_D \in \mathcal{R}^{m\times k}$
where $H_D =[h_{x_1},..., h_{x_m}]$ is the text sequence embedding, $H_Y =[h_{y_1},..., h_{y_n}]$ is the label sequence embedding and $M \in \mathcal{R}^{m\times n}$.
Considering the semantic information among consecutive words, we further generalize $M$ through non-linearity network. 
Specifically, for a text fragment of length $2r + 1$ centered at $i$, the local matrix block $M_{i-r;i+r}$ in $M$ measures the correlation for the label-phrase pairs.
To improve the effectiveness of the sparse regularization, we use CNN with ReLU activation in the hidden layers, and perform max-pooling and hyperbolic tangent sequentially in the function $\Omega$:
\begin{equation}\label{eq3}
\overrightarrow{c} = \Omega(M_{i-r;i+r}) \cdot H_D      
\end{equation}
Note that the final document representation $\overrightarrow{c}$ is generated by aggregation of word representations $H_D$, and weighted by the label-specific attention vector $\Omega(\cdot)$.
\subsubsection{Label Predication}
Once having the discriminative document representation, we build the multi-label text classifier via a fully connected layer that captures more fine-grained features from different regions of the document:
\begin{equation}\label{eq7}
\overrightarrow{p} = sigmoid(W_1\overrightarrow{c}^T + b_1)
\end{equation}
where $W_1\in \mathcal{R}^{n\times k}$ and $b_1 \in \mathcal{R}^{n}$. 
We use Binary Cross Entropy as the loss function for the multi-label classification problem:
\begin{equation}\label{eq8}
\mathcal{L}_{mlc} = -\sum_{i=1}^{n} [ q_i \ln p_i + (1 - q_i ) \ln (1 - p_i)]
\end{equation}
where $p_i=P(y_i|D)$ is the probability of $y_i$ predicted by the model, and $q_i\in\{0,1\}$ is the categorical information of $y_i$. 
We train the model by minimizing the cross-entropy error.
\subsection{Multi-Task Learning with Label Correlations}
\label{Subtask}
In this subsection, we introduce two auxiliary tasks, \emph{Pairwise Label Co-occurrence Prediction} (PLCP) and \emph{Conditional Label Co-occurrence Prediction} (CLCP), to explore the second-order and high-order label relationships, respectively.
\subsubsection{PLCP Task} 
Suppose that each document $D$ contains the corresponding label set $Y^{+}$ and the uncorresponding label set $Y^{-}$. In order to train the model to understand second-order label relationships, we propose a binarized label-pair prediction task named as PLCP that can be trivially generated from the multi-label classification corpus. 
The strategy of selecting label pairs for co-occurrence prediction is straightforward.
One part is sampled only from $Y^{+}$, which is marked as \emph{IsCo-occur}, and the other part is sampled from $Y^{+}$ and $Y^{-}$, respectively, which is marked as \emph{NotCo-occur}. 
To construct the manual training dataset, we empirically set the ratio of \emph{IsCo-occur} and \emph{NotCo-occur} to $\gamma$.
As Figure \ref{model_3task} shows, we concat the embedding of the two labels $[y_i, y_j]$ together as the input features.
% , and $p(y_j|D,y_i)$ denotes the probability of the co-occurrance of the two labels. 
The additional binary classifier is used to predict whether the state of the two labels is \emph{IsCo-occur} or \emph{NotCo-occur}. 
The loss function is as followed:
\begin{equation}
\mathcal{L}_{plcp} =  - [ q_{ij} \ln p_{ij} + (1 - q_{ij}) \ln (1 - p_{ij})]
\end{equation}
where $p_{ij}=p(y_j|D,y_i)$ denotes the output probability of the the co-occurrance of the label-pair, and q is the ground-truth where $q_{ij}=1$ means \emph{IsCo-occur} and $q_{ij}=0$ means \emph{NotCo-occur}.

\subsubsection{CLCP Task} 
To further learn the high-order label relationships, we propose the conditional label co-occurrence prediction (CLCP) task.
We first randomly pick $s$ labels from $Y^+$ to form $Y^G$, i.e. $Y^G \subseteq Y^+$, and then predict whether the remaining labels of $Y$ are relevant with them. Specifically, we introduce an additional position vector $E_Y = [e_{y_1} , ..., e_{y_n}]$, where $e_{y_i}=0$ indicates that $y_i$ at that position is the sampled label, i.e. $y_i \in Y^G$, and $e_{y_i}=1$ indicates $y_i \in Y - Y^G$. The average of the embedding of the zero-position labels $h_{y^G}$ is concatenated to each nonzero-position label embedding as the input features to predict whether each of remaining labels should be co-occurrence when knowing the sampled labels. In Figure \ref{model_3task}, $p(y_i|D,Y^G)$ denotes the probability of $y_i$ predicated by the additional sigmoid classifier. 
The loss for the classification is the sum of binary cross-entropy loss of each nonzero-position:
\begin{equation}
\mathcal{L}_{clcp}=  - \sum_{i=1}^{n-s} [ q_i \ln p_i + (1 - q_i ) \ln (1 - p_i)]
\end{equation}
where $q_i \in \{0,1\}$ is the ground-truth to denote whether the label $y_i$ should be co-occurrence with $Y^{G}$, and $p_i=p(y_i|D,Y^G)$ is the output probability of each masked label $y_i$.

\subsubsection{Training Objectives}
\label{loss}
The same inputs are first fed into the shared layers, then each sub-task module takes the contextualized token-level representations generated by joint embedding and produces a probability distribution for its own target labels. 
The overall loss can be calculated by:
\begin{equation}
\mathcal{L} = \mathcal{L}_{mlc} + \alpha \mathcal{L}_{plcp} + (1-\alpha) \mathcal{L}_{clcp}
\end{equation}
where $\alpha$ is a hyperparameter in (0, 1), $\mathcal{L}_{plcp}$ and $\mathcal{L}_{clcp}$ are task-specific Cross-Entropy loss for PLCP task and CLCP task, respectively. \footnote{We also implement it with three tasks together. Since the two auxiliary tasks have the similar goal, there is no performance gain.}

\section{Experimental Setup} \label{Experimental_Setup}
\subsection{Datasets}
We validate our proposed model on two multi-label text classification
datasets:
\textbf{Arxiv Academic Paper Dataset (AAPD)} \cite{yang2018sgm} collected 55,840 abstracts of papers in the field of computer science, which is organized into 54 related topics. In AAPD dataset, each paper is assigned multiple topics.
\textbf{Reuters Corpus Volume I (RCV1-V2)} \cite{lewis2004rcv1} is composed of 804,414 manually categorized newswire stories for research purposes. Each story in the dataset can be assigned multiple topics, and there are 103 topics in total.

Tabel \ref{datasets} shows statistics of datasets. Each dataset is divided into a training set, a validation set, and a test set. We followed the division of these two datasets by \citet{yang2018sgm}. 
\subsection{Evaluation Metrics}
Multi-label classification can be evaluated with a group of metrics, which capture different aspects of the task \cite{zhang2013review}.
Following the previous works \cite{yang2018sgm,order-free:aaai2020}, 
we adopt hamming loss, Micro/Macro-F1 scores as our main evaluation metrics. Micro/Macro-P and Micro/Macro-R are also reported to assist analysis.
% we adopt hamming loss, Micro- and Macro- average Precision, Recall, F1-Score as evaluation metrics. 
A Macro-average will treat all labels equally, whereas a Micro-average will weighted compute each label by its frequency.

\begin{table}[t]
\centering
\renewcommand{\arraystretch}{1}
\small
\setlength{\tabcolsep}{1.8mm}{
\begin{tabular}{lcccc}
\toprule[1.0pt]
\textbf{Dataset} & $|\mathcal{D}|$ & $|Y|$ & $\overline{|D_i|}$ & $\overline{|Y_i^+|}$\\
\hline
\textbf{AAPD} & 55,840 & 54 & 163.42 & 2.41 \\
\textbf{RCV1-V2} & 804,414 & 103 & 123.94 & 3.24 \\
\bottomrule[1.0pt]
\end{tabular}}
\caption{Statistics of datasets. Here, $|\mathcal{D}|$ and $|Y|$ denote the total number of documents and labels. $\overline{|D_i|}$ is the average length of all documents. $\overline{|Y_i^+|}$ means the average number of labels associated with the document.}\label{datasets}
\end{table}

\begin{table*}[t]
\centering
\renewcommand{\arraystretch}{1}
\small
\setlength{\tabcolsep}{0.45mm}{
\begin{tabular}{l|c| ccc| ccc|c| ccc| ccc}
\toprule[1.0pt]
~ & \multicolumn{7}{c|}{\texttt{AAPD dataset}} & \multicolumn{7}{c}{\texttt{RCV1-V2 dataset}}\\
\textbf{Algorithm} & \textbf{HL}$\downarrow$& \multicolumn{3}{|c|}{\textbf{Mi- P / R / F1}$\uparrow$} & \multicolumn{3}{|c|}{\textbf{Ma- P / R / F1}$\uparrow$} & \textbf{HL}$\downarrow$& \multicolumn{3}{|c|}{\textbf{Mi- P / R / F1}$\uparrow$} & \multicolumn{3}{|c|}{\textbf{Ma- P / R / F1}$\uparrow$} \\
\midrule[0.3pt]
\textbf{BR$^\dag$}\cite{boutella2004learning} & 0.0316& 64.4 /& 64.8 / & 64.6&-  &-  &- & 0.0086 & 90.4 / & 81.6 / & 85.8 & -   &-  &- \\
\textbf{CNN$^\dag$}\cite{kim2014convolutional} & 0.0256 & { }\textbf{84.9} / & 54.5 / & 66.4 &-  &-  &-  & 0.0089 & 92.2 / &79.8 / & 85.5  & -  & -  & -\\
\textbf{LEAM}\cite{wang2018joint}  & 0.0261& 76.5 /& 59.6 / & 67.0&52.4 / & 40.3 / & 45.6 & 0.0090&87.1/& 84.1 / & 85.6 & 69.5 / &65.8 / &67.6 \\
\textbf{LSAN}\cite{xiao2019label}  & 0.0242 & 77.7 / & 64.6 / & 70.6  &67.6 / & 47.2 / & 53.5  & 0.0075  & 91.3 / & 84.1 / & 87.5&74.9 / & 65.0 / & 68.4 \\
% 63.89/75.40	67.28
\textbf{BERT}\cite{devlin2019bert}  & 0.0224 & 78.6 / & 68.7 / & 73.4  &68.7 / & 52.1 / & 57.2 & 0.0073&\textbf{92.7} / & 83.2 / & 87.7 & 77.3 / & 61.9 / & 66.7 \\
\midrule[0.1pt]
\textbf{CC$^\dag$}\cite{read2011classifier} & 0.0306 & 65.7 / & 65.1 / & 65.4&-  &-  &-& 0.0087 & 88.7 / & 82.8 / & 85.7 & -   &-  &- \\
\textbf{SGM$^{\dag\clubsuit}$}\cite{yang2018sgm}  & 0.0251 & 74.6 / & 65.9 / & 69.9  &-   &-  &-& 0.0081&88.7 / & 85.0 / & 86.9& -  &-  &- \\
% & 0.713& 0.680& 0.681 \\
\textbf{Seq2Set$^{\dag \clubsuit}$}\cite{yang2019deep}  & 0.0247 & 73.9 / & 67.4 / & 70.5 &-   &-  &-& 0.0073 & 90.0 / & 85.8 / & 87.9  & -   &-  &-\\
\textbf{OCD$^{\dag \clubsuit}$}\cite{order-free:aaai2020}  & - &-  &-  &72.0&-  &-  &58.5& -   &-  &-  & - &-  &-  &-\\
\textbf{ML-R$^{\dag}$}\cite{multi-reasoner:nju2020}&0.0248&72.6 / &\textbf{71.8} / &72.2&-   &-  &-&0.0079&89.0 / &85.2 / &87.1&- &-  &-\\
\textbf{Seq2Seq$_{T}^{\clubsuit}$}\cite{nam2017maximizing} &0.0275& 69.8 / &68.2 / &69.0&56.2 / &53.7 / &	54.0&0.0074&88.5 / &\textbf{87.4} / &87.9&69.8 / &65.5 / &66.1\\
\textbf{SeqTag$_{Bert}$}&0.0238& 74.3/ &71.5 / &72.9&61.5 / &\textbf{57.5} / &58.5&0.0073&90.6 / &84.9 / &87.7&73.7 / &66.7 / &68.7\\
\midrule[0.1pt]
\textbf{LACO}  &0.0213&80.2 / & 69.6 / & 74.5&70.4 / & 54.0 / & 59.1  &0.0072&90.8 / & 85.6 / & 88.1&75.9 / & 66.6 / & 69.2\\
\textbf{LACO${+plcp}$} &\textbf{0.0212}&79.5 / & 70.8 / & \textbf{74.9}&68.4 / & 55.8 / & 59.9  & \textbf{0.0070}&90.8 / & 86.2 / & 88.4&76.1 / & 66.5 / & 69.2 \\
\textbf{LACO${+clcp}$} &0.0215&78.9 / &70.8 / & 74.7&\textbf{71.9} / & 56.6 / & \textbf{61.2}  &\textbf{0.0070}&90.6 / & 86.4 / &\textbf{88.5}&\textbf{77.6} / & \textbf{71.5} / & \textbf{73.1} \\
% \textbf{LACO${+plcp+clcp}$} &0.0216&79.1 / &70.4 / &74.5&70.8 / &56.1 / &60.8&\textbf{0.0070}&90.4 / & 86.7 / & \textbf{88.5} &\textbf{77.9} / & 69.3 / & 71.7 \\
\bottomrule[1pt]
\end{tabular}}
\caption{Predictive performance of each comparing algorithm on two datasets.
Hamming Loss (HL), Micro (Mi-) and Marco (Ma-) average Precision (P), Recall (R), F1-Score (F1) are used as evaluation metrics. 
The $\downarrow$ represents the lower score the better performance, and the $\uparrow$ is the opposite.
Models with $\dag$ denote for its results are quoted from previous papers.
Models with $\clubsuit$ are the Seq2Seq-based models.}\label{Microresult_v2}
\end{table*}
\subsection{Comparing Algorithms}
We adopt a various of methods as baselines, which can be divided into two groups according to whether the label correlations are considered.

The first group of approaches do not consider label correlations.
Binary Relevance (BR) \cite{boutella2004learning} amounts to independently training one binary classifier (linear SVM) for each label.
CNN \cite{kim2014convolutional} utilizes multiple convolution kernels to extract text features and then output the probability distribution over the label space.
LEAM \cite{wang2018joint} involves label embedding to obtain a more discriminative text representation in text classification.
LSAN \cite{xiao2019label} learns the label-specific text representation with the help of attention mechanisms.
We also implement a BERT \cite{devlin2019bert} classifier which first encodes a document into vector space and then outputs the probability for each label independently. 

The second group of methods consider label correlations.
Classifier Chains (CC) \cite{read2011classifier} transforms the MLTC problem into a chain of binary classification problems.
SGM \cite{yang2018sgm} proposes the Seq2Seq model with global embedding mechanism to capture label correlations. 
Seq2Set \cite{yang2019deep} presents deep reinforcement learning to improve the performance of the Seq2Seq model.
We also implement a Seq2Seq baseline with 12-layer transformer, named with Seq2Seq$_{T}$.
More recently, OCD \cite{order-free:aaai2020} proposes a framework including one encoder and two decoders for MLTC to alleviate exposure bias.
ML-Reasoner \cite{multi-reasoner:nju2020} employs a binary classifier to predict all labels simultaneously and applies a novel iterative reasoning mechanism.
Besides, we also provide another strong baseline:
SeqTag$_{Bert}$ transforms multi-label classification task into sequential tagging task, which first obtain embeddings of each label ($H_Y$ in Sec \ref{model}) by our shared encoder and then output a probability for each label sequentially by a BiLSTM-CRF model \cite{huang2015bilstmcrf}.

Results of BR, CNN, CC, SGM, Seq2Set, OCD and ML-R are cited in previous papers and results of other baselines are implemented by us. All algorithms follow the same data division.
\subsection{Experimental Setting}
We implement our model in Tensorflow and run on NVIDIA Tesla P40. We fine-tune models on the English base-uncased versions of BERT \footnote{https://github.com/google-research/bert}. The batch size is 32, and the maximum total input sequence length is 320. The window size of the additional layer is 10, and we set $\gamma$ as 0.5.
We use Adam \cite{Kingma2015AdamAM} with learning rate of 5e-5, and train the models by monitoring Micro-F1 score on the validation set and stopping the training if there is no increase in 50,000 consecutive steps.
\section{Results and Analysis}
In this section, we report the main experimental results of the baseline models and the proposed method on two text datasets. Besides, we analyze the performance on different frequency labels, and further evaluate whether our method effectively learns the label correlations through label-pair confidence distribution learning and label combination prediction. Finally, we give a detailed analysis of the convergence study which demonstrates the generalization ability of our method.
% \vspace{1mm}
\subsection{Experiment Results}
% \vspace{2mm}
\label{main results}
We report the experimental results of all comparing algorithms on two datasets in Table \ref{Microresult_v2}. 
The first block includes methods without learning label correlations. The second block is the methods considering label correlations, and the third block is our proposed {\laco} methods. 
As shown in Table \ref{Microresult_v2}, the {\laco}-based models outperform all baselines by a large margin in the main evaluation metrics.
The following observations can be made according to the results:
\begin{table}[t]
\centering
\renewcommand{\arraystretch}{1}
\small
\setlength{\tabcolsep}{1.5mm}{
\begin{tabular}{l|cc|cc}
\toprule[1.0pt]
~ & \multicolumn{2}{c|}{\texttt{AAPD }} & \multicolumn{2}{c}{\texttt{RCV1-V2 }}\\
\textbf{Model} & \textbf{HL}& \textbf{Mi-F1}  & \textbf{HL}& \textbf{Mi-F1}\\
\midrule[0.1pt]
\textbf{BERT}& 9.39e-09 &3.80e-10 &4.95e-04 & 3.67e-08 \\
\textbf{SeqTag$_{Bert}$} &7.76e-16 &1.86e-07 &4.95e-04&3.67e-08\\
% \midrule[0.1pt]
\bottomrule[1.0pt]
\end{tabular}}
\caption{Statistical analysis results. The P-values of {\laco} on significant test comparing with the two strong baselines BERT and SeqTag$_{Bert}$.
}\label{p_test}
\end{table}
\\\noindent $\bullet$  Our basic model {\laco} training only by the MLTC task significantly improves previous results on hamming loss and Micro-F1.
Specifically, on the AAPD dataset, comparing to Seq2Set which considers modeling the label correlations, our basic model decreases by $13.8\%$ on hamming loss and improves by $5.67\%$ on Micro-F1. Comparing with the label embedding method like LSAN, {\laco} achieves a reduction of $4.00\%$ hamming loss score and an improvement of $0.69\%$ Micro-F1 score on the RCV1-V2 dataset.
Also, BERT is still a strong baseline, which shows that obtaining a high-quality discriminative document representation is important for the MLTC task. 
Here, we train the {\laco} with 3 random seeds and calculate the mean and the standard deviation. We perform a significant test with {\laco} and the two strong baselines BERT and SeqTag$_{Bert}$ in Table \ref{p_test}.
Comparing with the two strong baseline models, all of the P-values of {\laco} are below the threshold (p $<$ 0.05), suggesting that the performance is statistically significant.
In addition, we implement Friedman test \cite{demvsar2006statistical} for hamming loss and Micro-F1 metrics. 
% when critical value = 2.8179 (#),
The Friedman statistics $F_F$ for hamming loss is 7.875 and for Micro-F1 is 6.125, when the corresponding critical value is 2.8179 ($\text{\#}$ comparing algorithms $k=12$, $\text{\#}$ datasets $N=2$). As a result, the null hypothesis of indistinguishable performance among the compared algorithms is clearly rejected at 0.05 significance level.
\\\noindent $\bullet$ Compared with SGM, Seq2Seq$_T$ does not achieve significantly improvements, but SeqTag$_{Bert}$ shows good performance based on the shared Transformer encoder between document and labels. Notably, the result of SeqTag$_{Bert}$ on Micro-F1 is comparable to BERT, but the result on Macro-F1 is observably higher. The above illustrates that label correlation information is more important for learning low frequency labels. 
\\\noindent $\bullet$  As for the results of the multi-task learning methods,  the two subtasks introduced by our method have a certain degree of improvement on the main metrics of the two datasets. Specifically, we observe that the PLCP task shows better performance and presents the best score of 74.9 on Micro-F1 for AAPD dataset, while the CLCP task presents the best performance on Micro-F1 for RCV1-V2 dataset as 88.5.
Furthermore, the proposed multi-task framework shows great improvements than the basic model {\laco} on Macro-F1, which demonstrates that the performance on low-frequency labels can be greatly improved through our label correlation guided subtasks.
There are more detailed analysis in Section \ref{Low-frequency Evaluation} and \ref{Problem of label combination overfitting}.
Notably, the CLCP task performs better on Marco-F1 by considering the high-order correlations. We also implement the experiment using the losses of three tasks together, while the combination of the two subtasks can not further improve the model performance comparing to {\laco}$_{+plcp}$ or {\laco}$_{+clcp}$, which we consider is due to the strong relevance between the two tasks.
\begin{table}[t]
\centering
\renewcommand{\arraystretch}{1}
\small
\setlength{\tabcolsep}{1mm}{
\begin{tabular}{l|ccc|ccc}
% \begin{tabular}{l|c|c|ccccccc|ccccccc}
\toprule[1.0pt]
~ & \multicolumn{3}{c|}{\texttt{AAPD }} & \multicolumn{3}{c}{\texttt{RCV1-V2 }}\\
\textbf{Model} & \textbf{HL}& \textbf{Mi-F} & \textbf{Ma-F}  & \textbf{HL}& \textbf{Mi-F} & \textbf{Ma-F}\\
\midrule[0.1pt]
\textbf{LACO} &0.0213&74.5&59.1&0.0072&88.1&69.2\\
\midrule[0.1pt]
\textbf{w/o JE }  &0.0237 &72.6 &57.7&0.0077&87.5&68.4\\
\textbf{w/o CA } &0.0220&73.5&58.4&0.0073&87.8&68.5\\
\textbf{w/o JE \text{\&} CA} &0.0224&73.4&57.2&0.0073&87.7&66.7\\
\bottomrule[1.0pt]
\end{tabular}}
\caption{Ablation over the proposed joint embedding (JE) and cross attention (CA) mechanisms using the {\laco} model on AAPD and RCV1-V2 datasets.
}\label{ablation}
\end{table}
\subsection{Ablation Study}
\label{Ablation Study}
In this section, we will demonstrate the effectiveness of two cores of the proposed {\laco} model, that is a document-label joint embedding (JE) mechanism, and a document-label cross attention (CA) mechanism.
Note that, the setting of \textbf{w/o JE \text{\&} CA} is equivalent to the BERT baseline in Tabel \ref{Microresult_v2}, which encode document only and predict the probability for each label based on [CLS].
In the \textbf{w/o JE} setting, document embedding is encoded by BERT while each label embedding is a learnable random initialized vector. Its label prediction layer is the same with {\laco}.
In the \mbox{\textbf{w/o CA}} setting, document and label embedding are obtained by BERT jointly, and probability for each label is predicted based on [CLS].
Tabel \ref{ablation} shows that JE and CA are both important to obtain a more discriminative text representation.
After removing JE and CA mechanism, the performance drops more in the AAPD dataset than RCV1-V2 dataset. We believe that is mainly due to the less of training instance in AAPD, which is more difficult to learn relevant features especially for those low-frequency labels.
\begin{figure}[t]
	\small
	  \centering
	   \includegraphics[width=7.7cm]{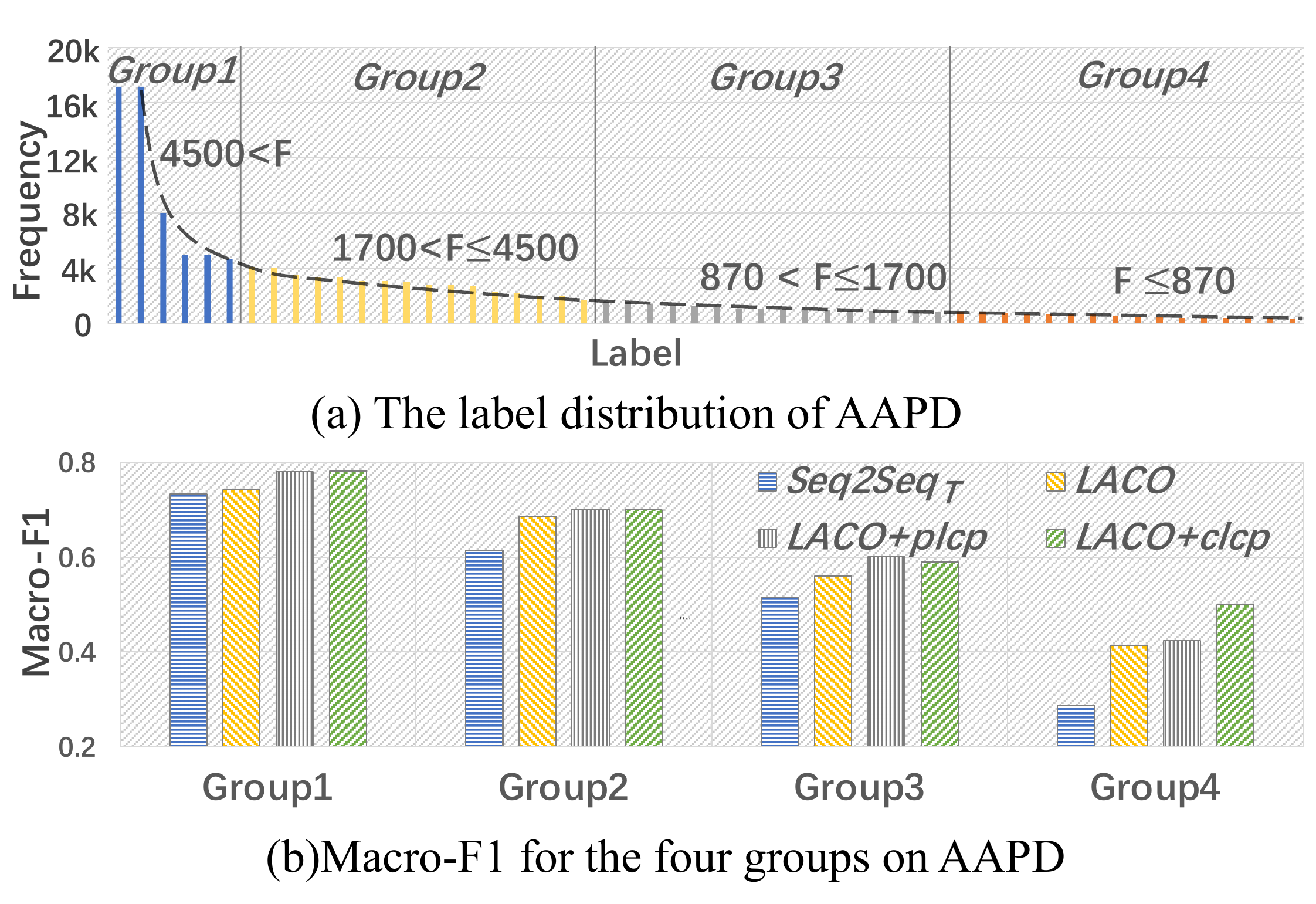}
	   \caption{Label classification performance on different frequency distributions. Subfigure(a) shows the label frequency distribution of each label on AAPD training set.
	   Subfigure(b) illustrates the Macro-F1 performance of different methods in the four groups.
	   }\label{fig_lowfreq}
\end{figure}
\subsection{Low-frequency Label Performance}
\label{Low-frequency Evaluation}
Figure \ref{fig_lowfreq}(a) illustrates the label frequency distribution on AAPD training set, which is a typical big-head-long-tail distribution.
We divide all the labels into four groups according to the frequency, the big-head group (Group1), the high-frequency group (Group2), the middle-frequency group (Group3), and the low-frequency group (Group4).
As shown in Figure \ref{fig_lowfreq}(b), we find the performance of all methods decreases with the label frequency of occurrence.
The performance gap between Seq2Seq$_T$ and {\laco}  based methods increases as the frequency decreases, especially in Group 4, {\laco}$_{+clcp}$ achieves a 74.5$\%$ improvement comparing to the Seq2Seq$_T$ model,
which demonstrates that the performance on low-frequency labels can be enhanced by the conditional label co-occurrence prediction task.
\subsection{Label Correlation Analysis}
The co-occurrence relationship between labels is one of the important aspects that can reflect label correlation.
In this experiment, we utilize the conditional probability $p(y_b|y_a)$ between label $y_a$ and $y_b$ to represent their dependency quantitatively.
Furthermore, we calculate the Conditional Kullback-Leibler Divergence of $p(y_b|y_a)$ to measure the ``distance" between model prediction distribution ($P^p$) and the ground-truth distribution on training/testing dataset ($P^g$). The score is calculate as:
\begin{equation}
\label{klequa}
\begin{aligned}
KL(P^{g}||P^{p}) &= \sum_{y_a,y_b\in Y}(p^{g}(y_b|y_a) log\frac{p^{g}(y_b|y_a)}{p^{p}(y_b|y_a)} \\
%P(l_b|l_a) &= \frac{\text{\#}(l_a, l_b)}{\text{\#}(l_a)}
p(y_b|y_a) &= \text{\#}(y_a, y_b) / \text{\#}(y_a)
\end{aligned}
\end{equation}
where $\text{\#}$ means the number of the single label or the label combination in the training/testing dataset. The KL-distances on the AAPD and RCV1-V2 datasets are shown in Table \ref{kl}. On the testing set settings, we can find that {\laco} has much better fitting ability for the dependency relationship between labels, especially after introducing the co-occurrence relationship prediction task.
The Seq2Seq$_T$ model achieves the lowest KL distance with training set on both AAPD and RCV1-V2 but achieve larger scores on the test set.
This conclusion further proves that the Seq2Seq-based model is prone to over-fitting label pairs during training. 
It should be emphasized that this KL distance just quantify how much interdependence between label pairs the model have learned, but it cannot directly measure the prediction accuracy of the model.

\subsection{Label Combination Diversity Analysis}
\label{Problem of label combination overfitting}
\begin{table}[t]
\centering
\renewcommand{\arraystretch}{1}
\small
\setlength{\tabcolsep}{3mm}{
\begin{tabular}{l|cc|cc}
% \begin{tabular}{l|c|c|ccccccc|ccccccc}
\toprule[1.0pt]
 & \multicolumn{2}{c|}{\texttt{AAPD }} & \multicolumn{2}{c}{\texttt{RCV1-V2 }}\\
Model & \textbf{train} & \textbf{test} & \textbf{train} & \textbf{test}\\
\hline
\textbf{Seq2Seq$_{T}$} &\textbf{1.27}& 1.30&0.08&0.94 \\
\textbf{SeqTag$_{Bert}$} &1.40&1.28&0.09 &0.95 \\
\textbf{LACO} &1.40 &1.27  &0.09&0.94\\
\textbf{LACO$_{+plcp}$}&1.35 &1.28 &0.08&\textbf{0.76}\\
\textbf{LACO$_{+clcp}$}&1.32& \textbf{1.10}&0.08&0.91\\
\bottomrule[1pt]
\end{tabular}}
\caption{
$KL(P^{g}||P^{p}$) for different models on AAPD and RCV1-V2 datasets. Note that $P^{g}$ is the ground truth distribution of datasets and  $P^{p}$ is the model distribution.
Smaller scores indicate that two distributions are closer.
}\label{kl}
\end{table}
Table \ref{overfit} shows the number of different predicted label combinations (\textbf{$C_{Test}$}) and subset accuracy (\textbf{$Acc$}), which is a strict metric that indicates the percentage of samples that have all their labels classified correctly.
Seq2Seq$_{T}$ produces fewer kinds of label combinations on the two datasets. As they tend to “remember” label combinations, the generated label sets are most alike, indicating a poor generalization ability to unseen label combinations. 
Because Seq2Seq$_{T}$ is conservative and only generates label combinations it has seen in the training set, it achieves high $Acc$ values, especially on RCV1-V2 dataset.
For our models, they produce more diverse label combinations while obtaining good $Acc$ since we do not regard multi-label classification as a sequence generation task that uses a decoder to model the relationship between labels. 
\begin{table}[t]
\centering
\renewcommand{\arraystretch}{1}
\small
\setlength{\tabcolsep}{2.5mm}{
\begin{tabular}{l|cc|cc}
\toprule[1pt]
~ & \multicolumn{2}{c|}{\texttt{AAPD }} & \multicolumn{2}{c}{\texttt{RCV1-V2 }}\\
\textbf{Model}  & \textbf{$C_{Test}$} & \textbf{$Acc$} &\textbf{$C_{Test}$}&\textbf{$Acc$}\\
% \textbf{$C_{Unseen~target}$}
\hline
\textbf{Ground~Truth}&392	&1.000	&278	&1.000\\
\textbf{Seq2Seq$_{T}$}&214	&0.392&87&0.669\\
\textbf{OCD$^\dag$} &302&0.403&-&-\\
% \textbf{BERT$_{leam}$}&277&149&	149	&67\\
\textbf{SeqTag$_{Bert}$}&289&0.410&187&0.637\\
\textbf{LACO}&315&0.425&241&0.642\\
\textbf{LACO$_{+plcp}$}&320&0.439&241&0.644\\
\textbf{LACO$_{+clcp}$}&321&0.427&239&0.660\\
\bottomrule[1pt]
\end{tabular}}
\caption{Statistics on the number of label combinations. $C_{Test}$ is the number of different predicted label combinations.
$Acc$ is the subset accuracy on the testing set.
}\label{overfit}
\end{table}
Instead, we learn the correlations among labels on the encoding side, and the scoring between labels does not interfere with each other, which leads to a higher probability of generating label combinations not seen during training than the Seq2Seq-based models.
\label{speed}
\begin{figure}[t]
	\small
	  \centering
	   \includegraphics[width=7.6cm]{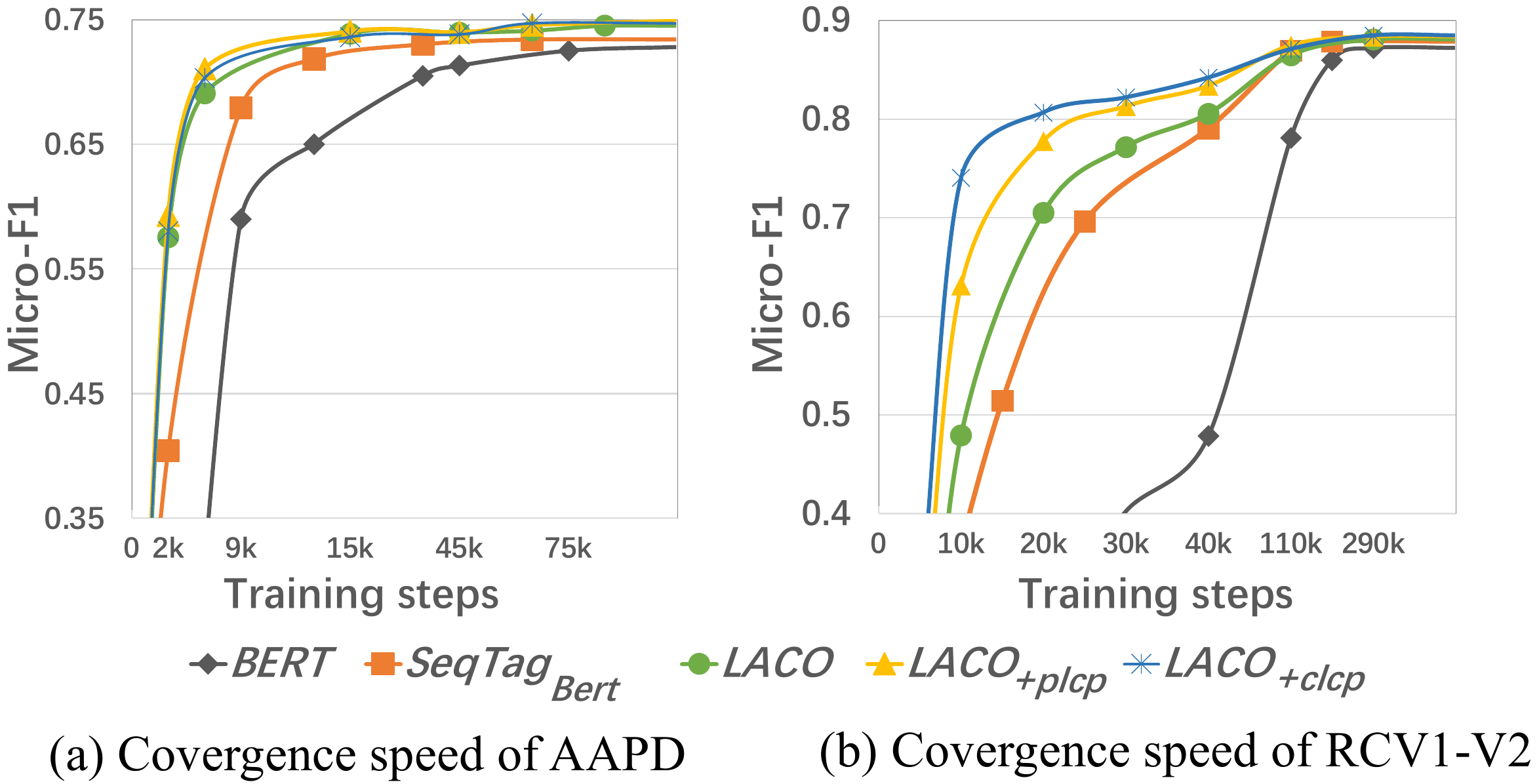}
	   \caption{The convergence speed of five BERT-based methods. The x-axis refers to the training steps, and the y-axis refers to the Micro-F1 score performance.}\label{fig_f1}
\end{figure}
\subsection{Coverage Speed}
The convergence speed of five BERT-based models are shown in Figure \ref{fig_f1}. 
Our basic model {\laco} outperforms other BERT-based models in terms of convergence speed, and the proposed multi-task mechanisms are able to enhance {\laco} to converge much faster. The main reason might be that the feature exchanging through multi-tasks accelerates the model to learn a more robust and common representation.
\section{Conclusions and Future Work}
In this paper, we propose a new method for MLTC based on document-label joint embedding and correlation aware multi-task learning. Experimental results show that our method outperforms competitive baselines by a large margin. Detailed analyses show the effectiveness of our proposed architecture using semantic connections between document-label and label-label, which helps to obtain a discriminative text representation. 
Furthermore, the multi-task framework shows strong capability on low-frequency label predicting and label correlation learning.

Considering the Extreme Multi-label Text Classification that contains an extremely large label set, {\laco} could be further exploited through scheduled label sampling, hierarchical label embedding strategy, and so on. We hope that further research could get clues from our work.

\section*{Acknowledgements}
We would like to thank the ACL reviewers for their valuable comments and Keqing He, Haoyan Liu, Zizhen Wang, Chenyang Liao and Rui Pan for their generous help and discussion. 
\bibliographystyle{acl_natbib}
\bibliography{anthology,acl2021}

\begin{thebibliography}{37}
\expandafter\ifx\csname natexlab\endcsname\relax\def\natexlab#1{#1}\fi

\bibitem[{Adhikari et~al.(2019)Adhikari, Ram, Tang, and
  Lin}]{adhikari2019docbert}
Ashutosh Adhikari, Achyudh Ram, Raphael Tang, and Jimmy Lin. 2019.
\newblock Docbert: Bert for document classification.
\newblock \emph{arXiv:1904.08398}.

\bibitem[{Bengio et~al.(2015)Bengio, Vinyals, Jaitly, and
  Shazeer}]{bengio2015scheduled}
Samy Bengio, Oriol Vinyals, Navdeep Jaitly, and Noam Shazeer. 2015.
\newblock Scheduled sampling for sequence prediction with recurrent neural
  networks.
\newblock In \emph{Proceedings of the 28th International Conference on Neural
  Information Processing Systems-Volume 1}, pages 1171--1179.

\bibitem[{Boutella et~al.(2004)Boutella, Luob, Shena, and
  Browna}]{boutella2004learning}
Matthew~R Boutella, Jiebo Luob, Xipeng Shena, and Christopher~M Browna. 2004.
\newblock Learning multi-label scene classification.
\newblock \emph{Pattern Recognition}, 37:1757--1771.

\bibitem[{Chen et~al.(2017)Chen, Ye, Xing, Chen, and
  Cambria}]{chen2017ensemble}
Guibin Chen, Deheng Ye, Zhenchang Xing, Jieshan Chen, and Erik Cambria. 2017.
\newblock Ensemble application of convolutional and recurrent neural networks
  for multi-label text categorization.
\newblock In \emph{2017 international joint conference on neural networks
  (IJCNN)}, pages 2377--2383. IEEE.

\bibitem[{Dem{\v{s}}ar(2006)}]{demvsar2006statistical}
Janez Dem{\v{s}}ar. 2006.
\newblock Statistical comparisons of classifiers over multiple data sets.
\newblock \emph{The Journal of Machine Learning Research}, 7:1--30.

\bibitem[{Devlin et~al.(2019)Devlin, Chang, Lee, and
  Toutanova}]{devlin2019bert}
Jacob Devlin, Ming-Wei Chang, Kenton Lee, and Kristina Toutanova. 2019.
\newblock Bert: Pre-training of deep bidirectional transformers for language
  understanding.
\newblock In \emph{Proceedings of the 2019 Conference of the North American
  Chapter of the Association for Computational Linguistics: Human Language
  Technologies, Volume 1 (Long and Short Papers)}, pages 4171--4186.

\bibitem[{Gibaja and Ventura(2015)}]{gibaja2015tutorial}
Eva Gibaja and Sebasti{\'a}n Ventura. 2015.
\newblock A tutorial on multilabel learning.
\newblock \emph{ACM Computing Surveys (CSUR)}, 47(3):1--38.

\bibitem[{Guo et~al.(2020)Guo, Han, Han, Huang, and Lu}]{guo2020label}
Biyang Guo, Songqiao Han, Xiao Han, Hailiang Huang, and Ting Lu. 2020.
\newblock Label confusion learning to enhance text classification models.
\newblock \emph{arXiv:2012.04987}.

\bibitem[{Huang et~al.(2015)Huang, Xu, and Yu}]{huang2015bilstmcrf}
Zhiheng Huang, Wei Xu, and Kai Yu. 2015.
\newblock Bidirectional lstm-crf models for sequence tagging.
\newblock \emph{arXiv:1508.01991}.

\bibitem[{Joachims(1998)}]{joachims1998text}
Thorsten Joachims. 1998.
\newblock Text categorization with support vector machines: Learning with many
  relevant features.
\newblock In \emph{European conference on machine learning}, pages 137--142.
  Springer.

\bibitem[{Kim(2014)}]{kim2014convolutional}
Yoon Kim. 2014.
\newblock Convolutional neural networks for sentence classification.
\newblock In \emph{Proceedings of the 2014 Conference on Empirical Methods in
  Natural Language Processing ({EMNLP})}, pages 1746--1751.

\bibitem[{Kingma and Ba(2015)}]{Kingma2015AdamAM}
Diederik~P. Kingma and Jimmy Ba. 2015.
\newblock Adam: {A} method for stochastic optimization.
\newblock In \emph{3rd International Conference on Learning Representations,
  {ICLR} 2015,San Diego, CA, USA, May 7-9, 2015, Conference Track Proceedings}.

\bibitem[{Kurata et~al.(2016)Kurata, Xiang, and Zhou}]{kurata2016improved}
Gakuto Kurata, Bing Xiang, and Bowen Zhou. 2016.
\newblock Improved neural network-based multi-label classification with better
  initialization leveraging label co-occurrence.
\newblock In \emph{Proceedings of the 2016 Conference of the North American
  Chapter of the Association for Computational Linguistics: Human Language
  Technologies}, pages 521--526.

\bibitem[{Lai et~al.(2015)Lai, Xu, Liu, and Zhao}]{lai2015recurrent}
Siwei Lai, Liheng Xu, Kang Liu, and Jun Zhao. 2015.
\newblock Recurrent convolutional neural networks for text classification.
\newblock In \emph{Proceedings of the AAAI Conference on Artificial
  Intelligence}, volume~29.

\bibitem[{Lewis et~al.(2004)Lewis, Yang, Rose, and Li}]{lewis2004rcv1}
David~D Lewis, Yiming Yang, Tony~G Rose, and Fan Li. 2004.
\newblock Rcv1: A new benchmark collection for text categorization research.
\newblock \emph{Journal of machine learning research}, 5(Apr):361--397.

\bibitem[{Liu et~al.(2017)Liu, Chang, Wu, and Yang}]{liu2017deep}
Jingzhou Liu, Wei-Cheng Chang, Yuexin Wu, and Yiming Yang. 2017.
\newblock Deep learning for extreme multi-label text classification.
\newblock In \emph{Proceedings of the 40th International ACM SIGIR Conference
  on Research and Development in Information Retrieval}, pages 115--124.

\bibitem[{Liu et~al.(2016)Liu, Qiu, and Huang}]{liu2016recurrent}
Pengfei Liu, Xipeng Qiu, and Xuanjing Huang. 2016.
\newblock Recurrent neural network for text classification with multi-task
  learning.
\newblock In \emph{Proceedings of the Twenty-Fifth International Joint
  Conference on Artificial Intelligence}, pages 2873--2879.

\bibitem[{Liu et~al.(2020)Liu, Shen, Wang, and Tsang}]{liu2020emerging}
Weiwei Liu, Xiaobo Shen, Haobo Wang, and Ivor~W Tsang. 2020.
\newblock The emerging trends of multi-label learning.
\newblock \emph{arXiv:2011.11197}.

\bibitem[{Liu and Tsang(2015)}]{liu2015optimality}
Weiwei Liu and Ivor~W Tsang. 2015.
\newblock On the optimality of classifier chain for multi-label classification.
\newblock In \emph{Proceedings of the 28th International Conference on Neural
  Information Processing Systems-Volume 1}, pages 712--720.

\bibitem[{Menon et~al.(2020)Menon, Jayasumana, Rawat, Jain, Veit, and
  Kumar}]{menon2020long}
Aditya~Krishna Menon, Sadeep Jayasumana, Ankit~Singh Rawat, Himanshu Jain,
  Andreas Veit, and Sanjiv Kumar. 2020.
\newblock Long-tail learning via logit adjustment.
\newblock \emph{arXiv:2007.07314}.

\bibitem[{Nam et~al.(2017)Nam, Menc{\'\i}a, Kim, and
  F{\"u}rnkranz}]{nam2017maximizing}
Jinseok Nam, Eneldo~Loza Menc{\'\i}a, Hyunwoo~J Kim, and Johannes
  F{\"u}rnkranz. 2017.
\newblock Maximizing subset accuracy with recurrent neural networks in
  multi-label classification.
\newblock In \emph{Advances in neural information processing systems}, pages
  5413--5423.

\bibitem[{Qin et~al.(2019)Qin, Li, Pavlu, and Aslam}]{qin2019adapting}
Kechen Qin, Cheng Li, Virgil Pavlu, and Javed Aslam. 2019.
\newblock Adapting rnn sequence prediction model to multi-label set prediction.
\newblock In \emph{Proceedings of the 2019 Conference of the North American
  Chapter of the Association for Computational Linguistics: Human Language
  Technologies, Volume 1 (Long and Short Papers)}, pages 3181--3190.

\bibitem[{Read et~al.(2011)Read, Pfahringer, Holmes, and
  Frank}]{read2011classifier}
Jesse Read, Bernhard Pfahringer, Geoff Holmes, and Eibe Frank. 2011.
\newblock Classifier chains for multi-label classification.
\newblock \emph{Machine learning}, 85(3):333.

\bibitem[{Tsai and Lee(2020)}]{order-free:aaai2020}
Che-Ping Tsai and Hung-Yi Lee. 2020.
\newblock Order-free learning alleviating exposure bias in multi-label
  classification.
\newblock In \emph{Proceedings of the AAAI Conference on Artificial
  Intelligence}, volume~34, pages 6038--6045.

\bibitem[{Tsoumakas and Katakis(2007)}]{tsoumakas2007multi}
Grigorios Tsoumakas and Ioannis Katakis. 2007.
\newblock Multi-label classification: An overview.
\newblock \emph{International Journal of Data Warehousing and Mining (IJDWM)},
  3(3):1--13.

\bibitem[{Tsoumakas et~al.(2009)Tsoumakas, Katakis, and
  Vlahavas}]{tsoumakas2009mining}
Grigorios Tsoumakas, Ioannis Katakis, and Ioannis Vlahavas. 2009.
\newblock Mining multi-label data.
\newblock In \emph{Data mining and knowledge discovery handbook}, pages
  667--685. Springer.

\bibitem[{Vinyals et~al.(2015)Vinyals, Bengio, and Kudlur}]{vinyals2015order}
Oriol Vinyals, Samy Bengio, and Manjunath Kudlur. 2015.
\newblock Order matters: Sequence to sequence for sets.

\bibitem[{Wang et~al.(2018)Wang, Li, Wang, Zhang, Shen, Zhang, Henao, and
  Carin}]{wang2018joint}
Guoyin Wang, Chunyuan Li, Wenlin Wang, Yizhe Zhang, Dinghan Shen, Xinyuan
  Zhang, Ricardo Henao, and Lawrence Carin. 2018.
\newblock Joint embedding of words and labels for text classification.
\newblock In \emph{Proceedings of the 56th Annual Meeting of the Association
  for Computational Linguistics (Volume 1: Long Papers)}, pages 2321--2331.

\bibitem[{Wang et~al.(2020)Wang, Ridley, Qu, Dai
  et~al.}]{multi-reasoner:nju2020}
Ran Wang, Robert Ridley, Weiguang Qu, Xinyu Dai, et~al. 2020.
\newblock A novel reasoning mechanism for multi-label text classification.
\newblock \emph{Information Processing \& Management}, 58(2):102441.

\bibitem[{Xiao et~al.(2019)Xiao, Huang, Chen, and Jing}]{xiao2019label}
Lin Xiao, Xin Huang, Boli Chen, and Liping Jing. 2019.
\newblock Label-specific document representation for multi-label text
  classification.
\newblock In \emph{Proceedings of the 2019 Conference on Empirical Methods in
  Natural Language Processing and the 9th International Joint Conference on
  Natural Language Processing (EMNLP-IJCNLP)}, pages 466--475.

\bibitem[{Xun et~al.(2020)Xun, Jha, Sun, and
  Zhang}]{correlation-network:kdd2020}
Guangxu Xun, Kishlay Jha, Jianhui Sun, and Aidong Zhang. 2020.
\newblock Correlation networks for extreme multi-label text classification.
\newblock In \emph{Proceedings of the 26th ACM SIGKDD International Conference
  on Knowledge Discovery \& Data Mining}, pages 1074--1082.

\bibitem[{Yang et~al.(2019)Yang, Luo, Ma, Lin, and Sun}]{yang2019deep}
Pengcheng Yang, Fuli Luo, Shuming Ma, Junyang Lin, and Xu~Sun. 2019.
\newblock A deep reinforced sequence-to-set model for multi-label
  classification.
\newblock In \emph{Proceedings of the 57th Annual Meeting of the Association
  for Computational Linguistics}, pages 5252--5258.

\bibitem[{Yang et~al.(2018)Yang, Sun, Li, Ma, Wu, and Wang}]{yang2018sgm}
Pengcheng Yang, Xu~Sun, Wei Li, Shuming Ma, Wei Wu, and Houfeng Wang. 2018.
\newblock Sgm: sequence generation model for multi-label classification.
\newblock \emph{In Proceedings of the 27th International Conference on
  Computational Linguistics}, page 3915–3926.

\bibitem[{Yang et~al.(2016)Yang, Yang, Dyer, He, Smola, and
  Hovy}]{yang2016hierarchical}
Zichao Yang, Diyi Yang, Chris Dyer, Xiaodong He, Alex Smola, and Eduard Hovy.
  2016.
\newblock Hierarchical attention networks for document classification.
\newblock In \emph{Proceedings of the 2016 conference of the North American
  chapter of the association for computational linguistics: human language
  technologies}, pages 1480--1489.

\bibitem[{You et~al.(2018)You, Zhang, Wang, Dai, Mamitsuka, and
  Zhu}]{you2018attentionxml}
Ronghui You, Zihan Zhang, Ziye Wang, Suyang Dai, Hiroshi Mamitsuka, and
  Shanfeng Zhu. 2018.
\newblock Attentionxml: Label tree-based attention-aware deep model for
  high-performance extreme multi-label text classification.
\newblock \emph{arXiv:1811.01727}.

\bibitem[{Zhang and Zhou(2014)}]{zhang2013review}
Min-Ling Zhang and Zhi-Hua Zhou. 2014.
\newblock A review on multi-label learning algorithms.
\newblock \emph{IEEE transactions on knowledge and data engineering},
  26(8):1819--1837.

\bibitem[{Zhao et~al.(2020)Zhao, Gao, Chen, and Wang}]{Seq2Seq+br}
Wei Zhao, Hui Gao, Shuhui Chen, and Nan Wang. 2020.
\newblock Generative multi-task learning for text classification.
\newblock \emph{IEEE Access}, 8:86380--86387.

\end{thebibliography}
%\appendix
\end{document}